\documentclass[letterpaper]{easychair}

\usepackage{natbib}
\usepackage{amsmath}
\usepackage{amssymb}
\usepackage{booktabs}
\usepackage{array}
\usepackage{enumitem}
\usepackage{graphicx}
\usepackage{indentfirst}
\usepackage{microtype}
\usepackage{url}

\setlist[itemize]{leftmargin=1.5em,nosep}
\hypersetup{
  pdfauthor={Junfeng Nie and Xiaohui Chen},
  pdftitle={DSFlow: A Neuro-Symbolic Data Workflow Agent with Statistical Diagnostic Guardrails},
  pdfsubject={Neuro-symbolic agents for data analysis},
  pdfkeywords={data agents, agent skills, neuro-symbolic AI, statistical guardrails, auditable workflows, data automation}
}

\newcommand{\accept}{\textsc{Accept}}
\newcommand{\override}{\textsc{Override}}

\makeatletter
\newenvironment{appendixfigure}{%
  \par\noindent\begin{minipage}{\textwidth}
  \centering
  \def\@captype{figure}%
}{%
  \end{minipage}\par
}
\makeatother

\title{DSFlow: A Neuro-Symbolic Data Workflow Agent with Statistical Diagnostic Guardrails}
\author{Junfeng Nie\inst{1,2} \and Xiaohui Chen\inst{1,2}}
\institute{AnyFluxion \and University of Southern California\\[3pt]
  \email{junfeng.nie@anyfluxion.com}, \email{xiaohui.chen@anyfluxion.com}}
\authorrunning{Junfeng Nie and Xiaohui Chen}
\titlerunning{DSFlow: a neuro-symbolic data workflow agent}

\begin{document}
\maketitle

\begingroup
  \renewcommand{\thefootnote}{}
  \footnotetext{This work was presented as a poster at the Chen Institute Symposium for AI Advancing Science and Society (AIAS+ 2026), held November 5--7, 2026, in San Francisco, California, USA.}
\endgroup

\begin{abstract}
Data science workflows often require tools that presume non-trivial technical knowledge. Large language models (LLMs) reduce this barrier, but a gap remains between flexible general-purpose agents and statistically controlled pipelines. We introduce \emph{DSFlow}, a LangGraph-based neuro-symbolic data-agent platform that combines reusable, LLM-selected skills with deterministic diagnostic guardrails. Skills provide extensible task knowledge. Symbolic validation can accept a proposed method, request clarification or replanning, override an unsuitable method, or supply a safe fallback. Regression and clustering case studies illustrate how DSFlow uses diagnostic evidence to correct method choices and documents those decisions in auditable reports.
\end{abstract}


\section{Introduction}
\label{sec:introduction}

Modern data science is increasingly mediated by heterogeneous computational workflows. A typical analysis pipeline requires data intake, cleaning, visualization, statistical testing, predictive modeling, diagnostic checking, evaluation, model comparison, and report generation. Each stage relies on different software libraries and methodological conventions, and each implicitly assumes that the user knows how to select appropriate tools, verify statistical assumptions, interpret outputs, and revise the analysis when the chosen method is inappropriate. This creates a persistent gap between high-level analytical intent and reliable executable data workflows: users can often express what they want to learn from data in natural language, but they may not know how to materialize that intent as a statistically rigorous pipeline.

Meanwhile, large language models (LLMs) have substantially reduced this barrier by enabling chat-based interfaces for code generation, tool use, reasoning, and summarization. ReAct introduced a general paradigm in which LLMs interleave reasoning with task-specific actions, allowing a model to plan, call tools, observe intermediate results, and update subsequent decisions~\citep{yao2022react}. Neuro-symbolic architectures such as MRKL further embed LLMs in modular systems that combine language understanding with specialized tools, external knowledge, and discrete reasoning modules~\citep{karpas2022mrkl}. Skill-based architectures extend this modular view by packaging procedural knowledge, code, and resources into reusable capabilities that can be loaded on demand~\citep{xu2026agentskills}. Together, these developments have shifted LLM applications from passive text generation toward interactive, tool-using, and workflow-oriented agents.

Data science is a natural domain for such agents. DS-Agent retrieves and adapts prior solutions from data-science competitions, Data Interpreter constructs executable workflow graphs, and DataClaw combines reasoning, memory, local execution, and pluggable skills~\citep{guo2024dsagent,hong2024datainterpreter,li2026dataclaw}. Recent surveys describe a rapidly growing landscape spanning data preparation, modeling, evaluation, visualization, deployment, and monitoring~\citep{chen2025dsagentsurvey,luo2026dataagents}. These systems can reduce the effort required to translate a high-level request into code and artifacts, making sophisticated workflows accessible to users who may not be expert programmers.

This progress also changes where analytical risk enters the workflow. When an assistant can generate and execute an entire pipeline, an unsuitable modeling choice may be propagated automatically from data preparation through evaluation and reporting. A coherent narrative or a successfully executed notebook does not by itself establish that the selected method matches the structure of the data. Reliable assistance therefore requires both broad procedural capability and a visible account of why consequential analytical decisions were made.

This concern also distinguishes DSFlow's objective from conventional pipeline automation. Automating model search or hyperparameter selection can improve predictive performance within a predefined task, but it does not determine whether the task, target, validation scheme, or model family is appropriate in the first place. DSFlow addresses this earlier control problem while still exposing the resulting models, diagnostics, and artifacts to the user.

This need becomes concrete at method selection. An agent may overlook violated assumptions or influential observations, optimize an unsuitable validation criterion, confuse exploratory visualization with confirmatory inference, or produce a plausible report unsupported by the data-generating structure. The appropriate workflow depends on the analytical task and on properties such as sample size, feature dimension, nonlinearity, dependence, anomaly structure, class balance, and geometry. Because users may not know which assumptions should have been checked, statistically consequential choices cannot safely depend on fluent language generation alone.

We introduce \emph{DSFlow}, a LangGraph-based neuro-symbolic platform for data analytics with statistical guardrails. Its design treats LLM reasoning, reusable skills, and symbolic rules as complementary. Reusable \texttt{SKILL.md} modules allow the LLM to interpret user intent and propose task-specific plans without hard-wiring every request. Deterministic validators ground plans in the uploaded schema, while executable diagnostics retain authority over statistically consequential choices. A guardrail can accept a valid skill preference, request clarification or replanning, override an unsuitable method, or supply a symbolic fallback before execution and reporting.

This work makes three contributions.
\begin{itemize}
  \item DSFlow integrates LLM-selected skills, schema validation, deterministic diagnostic routing, and execution checks in one workflow that records its decisions.
  \item A controlled regression and clustering study measures when the router retains or overrides a proposed method and compares downstream outcomes across routing policies.
  \item A separate prompt study measures task and schema grounding, first-pass plan validity, clarification, and LLM overhead for two instruction models.
\end{itemize}

We evaluate DSFlow first through a semi-synthetic regression case with injected influential records and a fully synthetic non-convex clustering case. In both settings, the system must diagnose why a plausible default is unsuitable and execute a supported alternative. Additional controlled experiments examine whether the same architecture preserves appropriate preferences, grounds natural-language requests, and remains stable under component and threshold variations. The broader aim is not to replace scientific judgment with a fixed catalogue of rules, but to provide an extensible agent architecture in which new capabilities can be added without surrendering consequential decisions to opaque language-model behavior. Scientific data analysis is a core stage of research and decision making; DSFlow is therefore designed not merely to automate code generation, but to make method selection and revision inspectable and statistically constrained.

\section{Related Work}
\label{sec:related_work}

Recent systems increasingly translate natural-language analytical goals into executable data workflows. LIDA decomposes visualization authoring into specialized stages, while Data-Copilot invokes pre-designed, compiler-validated interfaces~\citep{dibia2023lida,zhang2023datacopilot}. DS-Agent adapts prior solutions, Data Interpreter constructs executable workflow graphs, and Causal-Copilot routes among specialized analytical procedures~\citep{guo2024dsagent,hong2024datainterpreter,wang2025causalcopilot}. These systems establish the value of modular capabilities, structured execution, and natural-language access. DSFlow builds on this direction by giving data-derived diagnostics an explicit role in accepting, revising, or replacing a proposed method; this is a difference in emphasis, not a claim that other systems lack every form of validation.

Recent benchmarks also expose the difference between producing an answer and completing a reliable analytical workflow. InfiAgent-DABench evaluates executable questions derived from tabular data, BLADE evaluates open-ended scientific analyses against expert decisions, DataSciBench uses task-, function-, and code-level criteria, and Spider2-V evaluates professional data-science and engineering workflows in graphical environments~\citep{hu2024dabench,gu2024blade,zhang2025datascibench,cao2024spider2v}. MLE-Bench extends this perspective to 75 end-to-end machine-learning competitions~\citep{chan2025mlebench}. These benchmarks provide breadth across tasks and environments. Our study is complementary: it uses controlled cases with known expectations to inspect diagnostic decisions and the resulting executed workflow.

\section{System Design}
\label{sec:system}

DSFlow uses LangGraph to maintain a shared typed state across parsing, skill selection, planning, validation, diagnostics, routing, execution, reporting, and memory update. The LLM parses intent, selects reusable \texttt{SKILL.md} capabilities, and proposes a plan containing the task, target, features, model preference, metric, and constraints. A deterministic plan guardrail verifies supported vocabularies and grounds requested fields in the uploaded schema. Missing or conflicting information produces a clarification request; invalid model or metric choices trigger replanning and then a symbolic fallback if retries fail. The plan is therefore a structured proposal rather than direct executable authority.

\begin{figure}[tb]
\centering
\includegraphics[width=\textwidth]{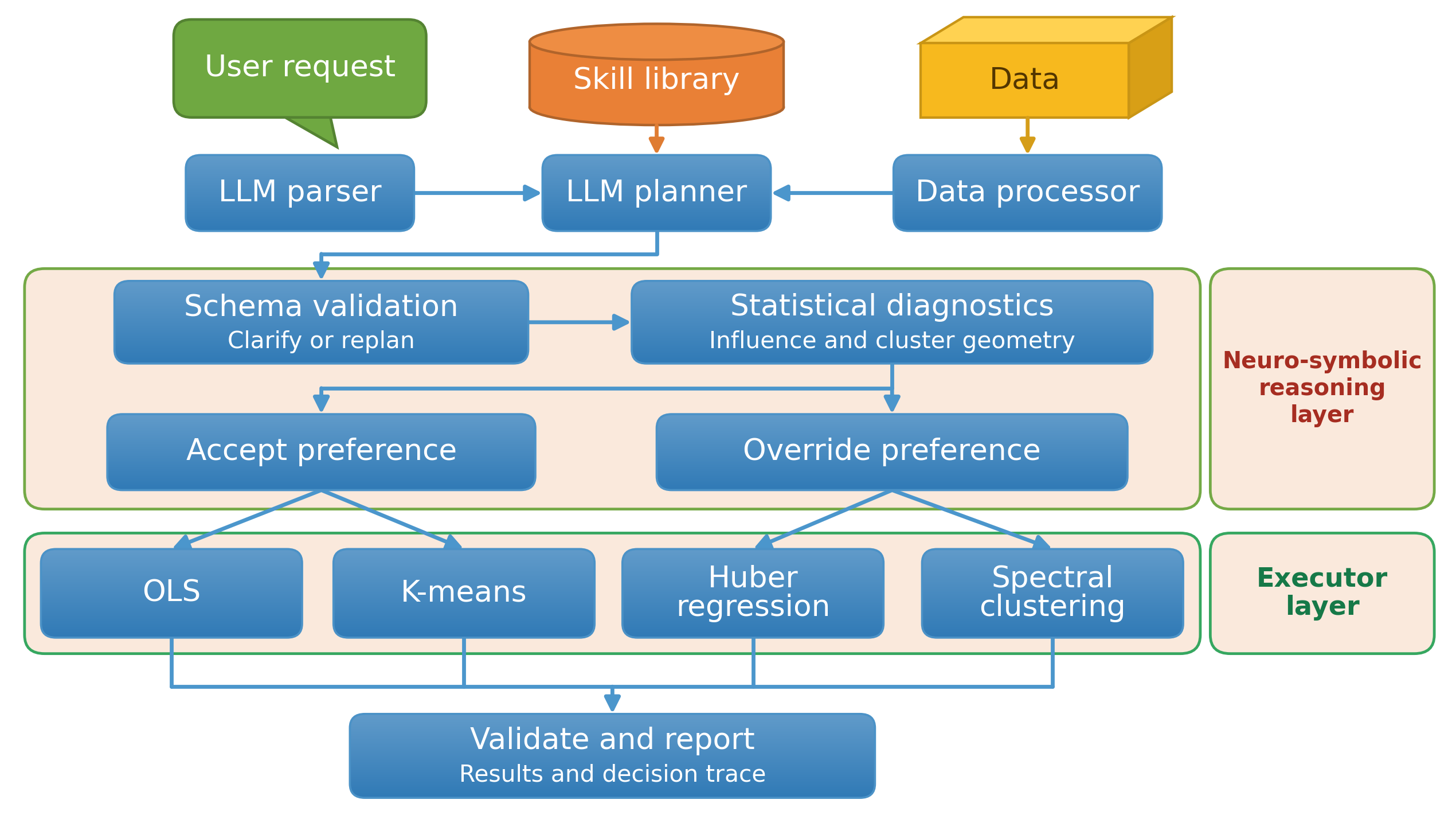}
\caption{DSFlow workflow for the evaluated regression and clustering paths. The LLM proposes a plan. Schema validation can request clarification or replanning, and diagnostics determine whether to retain or override the proposed method. The examples pair ordinary least squares (OLS) with Huber robust regression and $K$-means with spectral clustering. A deterministic stage checks the executor's outputs and formats the report. Classification is supported in the architecture and outside the scope of the evaluation.}
\label{fig:dsflow-architecture}
\end{figure}

Figure~\ref{fig:dsflow-architecture} shows the regression and clustering paths evaluated in this paper. The planner also supports classification. Task-specific checkers inspect properties such as nonlinearity, heteroscedasticity, class imbalance, dimensionality, influence, and data geometry to inform method selection. The executor returns typed numerical results, diagnostics, visualizations, and structured artifacts.

An execution guardrail checks structural validity before the reporter formats computed facts, caveats, and artifact references. The reporter does not introduce a new target, method, or statistical conclusion after execution. Each run records the grounded task, requested and executed methods, diagnostic evidence, router decision, clarification state, and output artifacts, producing an audit trace that can be inspected or replayed under alternative routing policies.

The platform also provides session memory and optional semantic retrieval. Request-scoped telemetry records operational LLM metadata such as model, stage, token count, and latency, while product session and episodic-memory stores are managed separately. Semantic retrieval was disabled in the reported experiments. This paper evaluates regression and clustering in depth; classification remains in the architectural scope only.

\section{Evaluation}
\label{sec:evaluation}

We use two original case studies to illustrate the end-to-end workflow. The controlled routing study then isolates method selection by supplying predefined plans. A separate prompt study evaluates the language interface, and component and threshold experiments examine specific implementation choices.

\subsection{Original comparative case study}

The original comparison used two settings in which a plausible default method fails unless the system diagnoses the problem and executes a correction. The regression case augments 5,000 USA Housing records~\citep{vedavyasvUSAHousing}: after removing the address field, two influential records are added, yielding 5,002 rows. The clustering dataset is fully synthetic and contains a central disc, an inner ring, and an outer ring with sizes $250/250/500$, creating a non-convex geometry for which centroid-based partitions can appear numerically plausible while misrepresenting the generating structure.

DSFlow, Manus 1.6 Lite, Gemini 3.1 Flash-Lite, and ChatGPT 5.5 Instant each received the same CSV and one-shot task three times in fresh conversations, without follow-up correction. The prompt requested an end-to-end analysis rather than prescribing a model, so each system had to inspect the data, choose and execute a method, evaluate the result, and communicate limitations. One author and one independent computer-science graduate student scored a seven-item, method-neutral rubric independently; suggested but unexecuted steps received no credit. Of 168 criterion judgments, 91.1\% agreed ($\kappa=0.82$) before joint adjudication.

For regression, the rubric covered influence detection, identification and handling of the injected records, predictive validation, quantified associations, interpretive limits, and visualization. For clustering, it covered the number and geometry of groups, use of a non-convex method, component sizes, structural validation, visualization, and avoidance of unsupported causal claims. These criteria evaluate completed analytical behavior rather than writing style or agreement with a particular software implementation.

The two cases isolate complementary analytical failures. In regression, an ordinary model remains executable but its estimates are sensitive to a few observations, so success requires influence diagnostics and an appropriate robust response. In clustering, a familiar internal criterion can favor a visually coherent but topologically incorrect partition, so success requires structural validation and execution of a non-centroid alternative. A system was therefore not rewarded merely for mentioning a problem or suggesting another method in prose; the correction had to appear in the executed analysis and its reported evidence.

Table~\ref{tab:original-comparison} summarizes the adjudicated comparison. Exact prompts, run-level scores, and criterion-level agreement are provided in the appendix.

\begin{table}[htbp]
\centering
\small
\caption{Original descriptive case-study scores (maximum 7 per run), reported as mean $\pm$ population SD over three runs.}
\label{tab:original-comparison}
\begin{tabular}{lccc}
\toprule
System & Regression & Clustering & Criteria met \\
\midrule
DSFlow & $7.00\pm0.00$ & $7.00\pm0.00$ & 42/42 \\
Manus & $2.67\pm0.47$ & $3.67\pm1.89$ & 19/42 \\
Gemini & $2.00\pm0.00$ & $1.33\pm0.47$ & 10/42 \\
ChatGPT & $1.00\pm0.00$ & $1.67\pm0.47$ & 8/42 \\
\bottomrule
\end{tabular}
\end{table}

\paragraph{Contaminated regression.}
DSFlow retained the injected extremes, computed Cook's distance~\citep{cook1977influential}, leverage, and externally studentized residuals on the training partition, and selected Huber robust regression~\citep{huber1964robust} through training-only validation. Ordinary least squares (OLS) fitted to all training rows had a holdout root mean squared error (RMSE) of 127,139, whereas robust regression retained all rows and reduced RMSE to 100,131. OLS after excluding the three most influential training rows reached 100,122, showing that robust fitting closely matched deletion-based OLS without treating screening as proof that a record should be discarded. The general-purpose assistants usually identified broad predictor rankings, but their handling of contamination was incomplete. Manus removed invalid or interquartile-range-flagged rows and used ordinary linear or random-forest analyses; Gemini filtered extreme observations before correlation or ordinary regression; and ChatGPT relied on random-forest feature importance while only suggesting robust analysis in some runs. None consistently combined influence diagnostics, robust estimation, uncertainty, and an isolated holdout evaluation.

\begin{figure}[tb]
\centering
\includegraphics[width=0.92\textwidth]{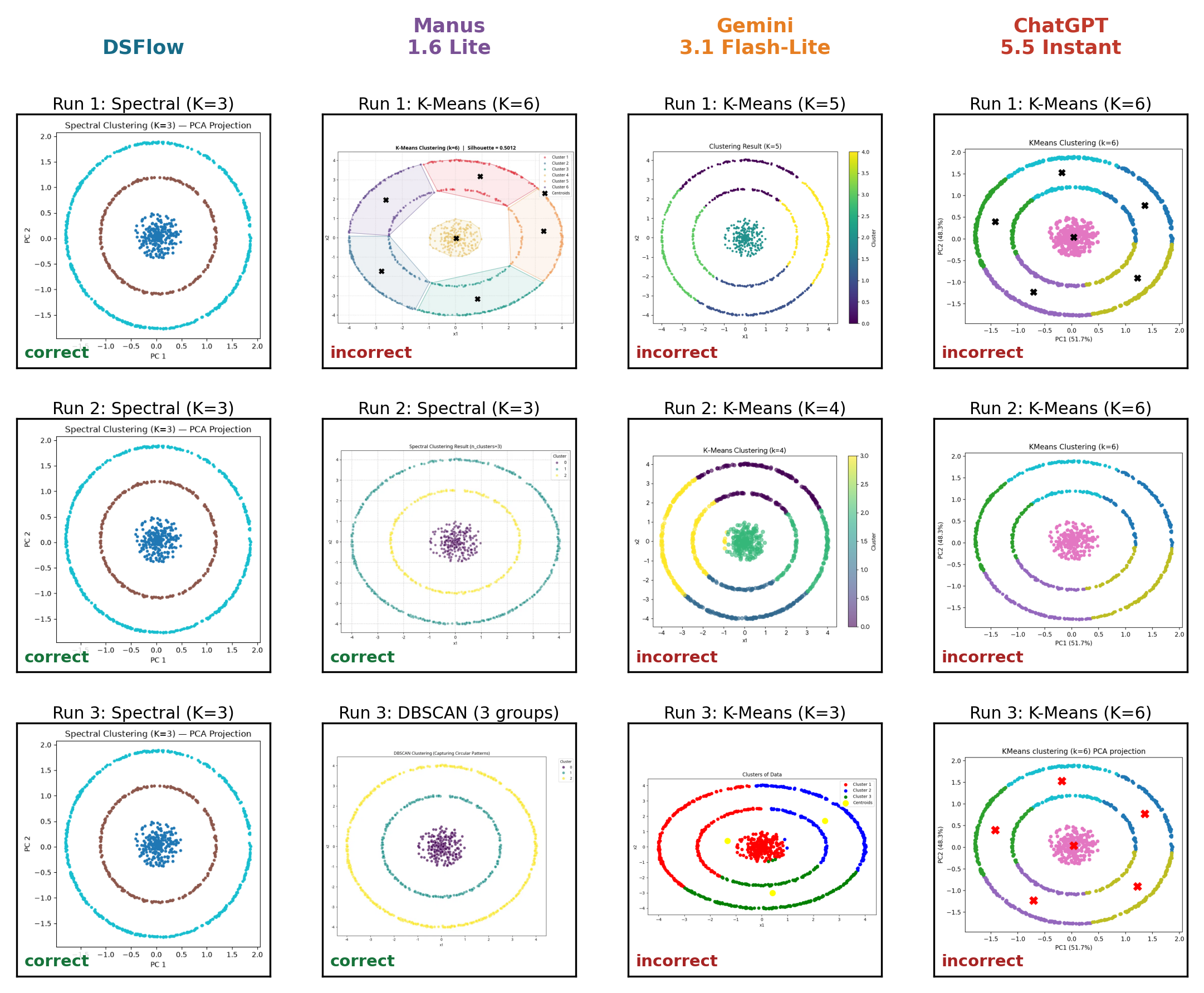}
\caption{Original product outputs for the non-convex clustering case across three executions per system. Point colors denote the cluster labels assigned by each system and need not correspond across panels. The label \emph{correct} denotes recovery of the central disc and both rings. Centroid solutions divide rings into angular sectors.}
\label{fig:clustering-runs}
\end{figure}

\paragraph{Non-convex clustering.}
Euclidean silhouette~\citep{rousseeuw1987silhouettes} favored angular partitions of the rings: $K$-means attained approximately 0.50 near six clusters even though the labels mixed inner- and outer-ring segments. DSFlow instead detected the same three connected components across local-neighbor values from 6 through 20 and routed the stable manifold structure to spectral clustering~\citep{ng2002spectral}; all three executions recovered the central disc and two rings. This geometry is a standard challenge for centroid-based manifold clustering~\citep{CHEN2021303}. Manus recovered the geometry in two of three runs using spectral clustering or DBSCAN, although neither successful report included both component sizes and a parameter-stability analysis. Gemini used $K$-means in all three runs, selecting between three and five clusters, and ChatGPT used a six-cluster $K$-means solution in every run. Two ChatGPT reports correctly described the non-convex limitation and proposed spectral or density-based alternatives, but did not execute the correction. Figure~\ref{fig:clustering-runs} shows the original product outputs rather than standardized reconstructions, so the panels preserve each system's own method choice and presentation.

\subsection{Selective statistical routing}

We supplied predefined plans to the deterministic core and evaluated four prespecified conditions at 20 seeds each. R-\accept{} uses a well-specified linear process with an OLS preference. R-\override{} replaces two rows with influential observations and expects a robust fit. C-\accept{} uses three isotropic Gaussian blobs with a $K$-means preference. C-\override{} uses a disc and two rings and expects spectral clustering. All policies were scored on whether the completed run executed the expected method. The selective router also had to record the expected accept or override decision.

\begin{table}[htbp]
\centering
\small
\caption{Final route accuracy across four controlled conditions.}
\label{tab:study-a}
\begin{tabular}{lccccc}
\toprule
Policy & R-A & R-O & C-A & C-O & Total \\
\midrule
Selective router & 18/20 & 19/20 & 20/20 & 20/20 & 77/80 \\
No override & 20/20 & 0/20 & 20/20 & 0/20 & 40/80 \\
Constant default & 20/20 & 0/20 & 20/20 & 0/20 & 40/80 \\
Constant complex & 0/20 & 20/20 & 0/20 & 20/20 & 40/80 \\
\bottomrule
\end{tabular}
\end{table}

The clustering router checks agreement between the stable local graph and the centroid partition before overriding a centroid preference. After correcting this agreement check, the selective router matched the expected route in 77/80 instances (96.25\%) on the primary set (Table~\ref{tab:study-a}). The constant-default, no-override, and constant-complex policies each matched 40/80. We then froze the corrected implementation and ran it on 40 additional instances generated from seeds 20--29 using the same four conditions. It matched 39/40 routes. All misses in both sets occurred in regression.

On the primary set, the selective router and the constant-complex policy both achieved a median adjusted Rand index (ARI) of 1.00 over 40 clustering instances and a median holdout RMSE of about 1.013 over 40 regression instances. In these conditions, the selective router retained appropriate simpler preferences while attaining similar median downstream performance to the constant-complex policy. Appendix~\ref{app:protocol} summarizes the generators, decision thresholds, and paired comparisons.

\subsection{Language interface and overhead}

The language interface was evaluated with 20 executable paraphrases and five ambiguity or guardrail prompts. Each was run once from fresh state with a 70B and an 8B Workers AI instruction model at temperature zero, yielding 50 model--prompt pairs. The endpoints, exact prompts, eligibility rules, and pricing snapshot are reported in Appendix~\ref{app:prompts}.

\begin{table}[htbp]
\centering
\scriptsize
\setlength{\tabcolsep}{2.3pt}
\caption{Language-interface results. Plan is first-pass validity among requests that reached planning. Clar. R is clarification recall on the five ambiguity and guardrail prompts.}
\label{tab:study-b}
\begin{tabular}{>{\raggedright\arraybackslash}p{0.31\textwidth}rrrrrr}
\toprule
Model & Task & Target & Features & Plan & Clar. R & Fails \\
\midrule
Llama 3.3 70B & 24/25 & 12/12 & 22/22 & 21/21 & 4/4 & 0 \\
Llama 3.1 8B & 24/25 & 12/12 & 22/22 & 21/21 & 4/4 & 0 \\
\bottomrule
\end{tabular}
\end{table}

Both model configurations produced 24/25 exact task-label matches, grounded the target and features on all eligible prompts, and produced valid first-pass plans on all 21 prompts that reached planning (Table~\ref{tab:study-b}). Each configuration requested clarification on all four expected cases and produced no false clarification on the single negative case. The only task-label mismatch was A3, where the scorer compared \texttt{UNKNOWN} to the reference value \texttt{null}. The system correctly requested clarification before any LLM call on that prompt. No run failed.

The median was two LLM calls per request. In the 25-record scope, the median total LLM latency per request was 2.83 seconds for 70B and 9.72 seconds for 8B, including zero-call requests. For records with reported usage, median input/output tokens and estimated cost were 981/140 and USD 0.000610 for 70B, and 933/167 and USD 0.000189 for 8B. Reporting added no LLM call.

\subsection{Components, thresholds, and baseline scope}

Table~\ref{tab:components} compares six component configurations on the original cases. The full core, oracle plan, and constant-complex policy produced the same robust-regression and spectral-clustering outcomes; disabling override increased regression RMSE by 26.97\% and replaced the radial components with six $K$-means sectors. The threshold results in Tables~\ref{tab:thresholds} and~\ref{tab:clustering-thresholds} show that low regression cutoffs reduce route selectivity, whereas all tested clustering cutoffs separate the controlled accept and override conditions. A time-bounded, zero-modification DS-Agent preflight exposed no task-equivalent unlabeled clustering workflow, so no score is assigned rather than constructing an outcome-specific adaptation.

\section{Discussion and Limitations}
\label{sec:discussion}

The broader design goal of DSFlow is to balance extensibility and trust. Skills allow procedural knowledge, code, and domain resources to grow without turning the system into a monolithic rule base, while LLMs provide a natural interface across heterogeneous tasks. Their flexibility also makes their choices difficult to trust at high-consequence decision points. Deterministic diagnostics are less universal, but they make evidence and transitions testable. DSFlow therefore combines them asymmetrically: skills propose, symbolic diagnostics decide, and guardrails can ask the LLM or user for a better-specified plan before execution.

The original comparative cases provide the clearest illustration of this design. General-purpose assistants sometimes recognized an outlier or non-convex geometry but did not consistently execute the corresponding method revision. DSFlow connected that evidence to robust regression and spectral clustering inside the workflow. The routing study further shows that the same architecture can retain a simpler preference when its assumptions remain compatible, rather than always selecting the most elaborate method. Language-interface and component results place LLM calls at intent and plan construction while leaving grounded numerical outcomes to the deterministic analytical core. The resulting trace links the request, schema, diagnostics, executed branch, and report for analyst review.

The hosted-product comparison covers two cases chosen to exercise diagnostic overrides. DSFlow's three executions per case recomputed the same fixed-seed workflow. One author participated in scoring and adjudication, and the hosted-product results reflect the evaluated versions and settings. The controlled router study covers four conditions, and the additional seeds test repeatability within those generators. Transfer to other data distributions, methods, and task families remains an open question. The interface study is limited to the 25 reported prompts, and its latency and cost depend on the dated hosted endpoints. Classification remains outside the evaluations.

\section{Conclusion}

DSFlow combines language-model planning with explicit checks on method selection and execution. In the evaluated regression and clustering conditions, it retained appropriate preferences or replaced them when diagnostics supported a change. The resulting decision trace connects the request, diagnostic evidence, executed method, and reported results.

\section{Reproducibility and Artifact Availability}
\label{sec:availability}

The appendices provide generator specifications, seeds, routing thresholds, prompts, component configurations, scoring rules, aggregate results, inter-rater agreement, and representative outputs. The implementation, generator scripts, manifests, and run records are retained in a private project repository.

\bibliographystyle{unsrt}
\bibliography{references}

@article{CHEN2021303,
  author={Chen, Xiaohui and Yang, Yun},
  title={Diffusion {K}-means clustering on manifolds: Provable exact recovery via semidefinite relaxations},
  journal={Applied and Computational Harmonic Analysis},
  volume={52},
  pages={303--347},
  year={2021},
  doi={10.1016/j.acha.2020.03.002}
}

@article{yao2022react,
  title={{ReAct}: Synergizing Reasoning and Acting in Language Models},
  author={Yao, Shunyu and Zhao, Jeffrey and Yu, Dian and Du, Nan and Shafran, Izhak and Narasimhan, Karthik and Cao, Yuan},
  journal={arXiv preprint arXiv:2210.03629},
  year={2022}
}

@article{karpas2022mrkl,
  title={{MRKL} Systems: A modular, neuro-symbolic architecture that combines large language models, external knowledge sources and discrete reasoning},
  author={Karpas, Ehud and Abend, Omri and Belinkov, Yonatan and Lenz, Barak and Lieber, Opher and Ratner, Nir and Shoham, Yoav and Bata, Hofit and Levine, Yoav and Leyton-Brown, Kevin and others},
  journal={arXiv preprint arXiv:2205.00445},
  year={2022}
}

@article{guo2024dsagent,
  title={{DS-Agent}: Automated Data Science by Empowering Large Language Models with Case-Based Reasoning},
  author={Guo, Siyuan and Deng, Cheng and Wen, Ying and Chen, Hechang and Chang, Yi and Wang, Jun},
  journal={arXiv preprint arXiv:2402.17453},
  year={2024}
}

@article{hong2024datainterpreter,
  title={Data Interpreter: An LLM Agent For Data Science},
  author={Hong, Sirui and Lin, Yizhang and Liu, Bang and Liu, Bangbang and Wu, Binhao and Zhang, Ceyao and Wei, Chenxing and Li, Danyang and Chen, Jiaqi and Zhang, Jiayi and others},
  journal={arXiv preprint arXiv:2402.18679},
  year={2024}
}

@article{li2026dataclaw,
  title={{DataClaw}: An Autonomous Data Agent with Instant Messaging Integration},
  author={Li, Huahang and Hu, Wentao and Wan, Zhuoyue and Zhang, Chen Jason and Li, Haoyang and Wei, Xiaoyong},
  journal={arXiv preprint arXiv:2604.24067},
  year={2026}
}

@inproceedings{dibia2023lida,
  title={{LIDA}: A Tool for Automatic Generation of Grammar-Agnostic Visualizations and Infographics using Large Language Models},
  author={Dibia, Victor},
  booktitle={Proceedings of the 61st Annual Meeting of the Association for Computational Linguistics (Volume 3: System Demonstrations)},
  pages={113--126},
  year={2023},
  doi={10.18653/v1/2023.acl-demo.11},
  url={https://aclanthology.org/2023.acl-demo.11/}
}

@article{zhang2023datacopilot,
  title={{Data-Copilot}: Bridging Billions of Data and Humans with Autonomous Workflow},
  author={Zhang, Wenqi and Shen, Yongliang and Tan, Zeqi and Hou, Guiyang and Lu, Weiming and Zhuang, Yueting},
  journal={arXiv preprint arXiv:2306.07209},
  year={2023},
  url={https://arxiv.org/abs/2306.07209}
}

@article{hu2024dabench,
  title={{InfiAgent-DABench}: Evaluating Agents on Data Analysis Tasks},
  author={Hu, Xueyu and Zhao, Ziyu and Wei, Shuang and Chai, Ziwei and Ma, Qianli and Wang, Guoyin and Wang, Xuwu and Su, Jing and Xu, Jingjing and Zhu, Ming and others},
  journal={arXiv preprint arXiv:2401.05507},
  year={2024},
  url={https://arxiv.org/abs/2401.05507}
}

@article{gu2024blade,
  title={{BLADE}: Benchmarking Language Model Agents for Data-Driven Science},
  author={Gu, Ken and Shang, Ruoxi and Jiang, Ruien and Kuang, Keying and Lin, Richard-John and Lyu, Donghe and Mao, Yue and Pan, Youran and Wu, Teng and Yu, Jiaqian and others},
  journal={arXiv preprint arXiv:2408.09667},
  year={2024},
  url={https://arxiv.org/abs/2408.09667}
}

@article{zhang2025datascibench,
  title={{DataSciBench}: An LLM Agent Benchmark for Data Science},
  author={Zhang, Dan and Zhoubian, Sining and Cai, Min and Li, Fengzu and Yang, Lekang and Wang, Wei and Dong, Tianjiao and Hu, Ziniu and Tang, Jie and Yue, Yisong},
  journal={arXiv preprint arXiv:2502.13897},
  year={2025},
  url={https://arxiv.org/abs/2502.13897}
}

@inproceedings{chan2025mlebench,
  title={{MLE-Bench}: Evaluating Machine Learning Agents on Machine Learning Engineering},
  author={Chan, Jun Shern and Chowdhury, Neil and Jaffe, Oliver and Aung, James and Sherburn, Dane and Mays, Evan and Starace, Giulio and Liu, Kevin and Maksin, Leon and Patwardhan, Tejal and Weng, Lilian and Madry, Aleksander},
  booktitle={International Conference on Learning Representations},
  year={2025},
  url={https://openreview.net/forum?id=6s5uXNWGIh}
}

@article{chen2025dsagentsurvey,
  title={Large Language Model-based Data Science Agent: A Survey},
  author={Chen, Ke and Wang, Peiran and Yu, Yaoning and Zhan, Xianyang and Wang, Haohan},
  journal={arXiv preprint arXiv:2508.02744},
  year={2025},
  doi={10.48550/arXiv.2508.02744},
  url={https://arxiv.org/abs/2508.02744}
}

@inproceedings{cao2024spider2v,
  title={{Spider2-V}: How Far Are Multimodal Agents From Automating Data Science and Engineering Workflows?},
  author={Cao, Ruisheng and Lei, Fangyu and Wu, Haoyuan and Chen, Jixuan and Fu, Yeqiao and Gao, Hongcheng and Xiong, Xinzhuang and Zhang, Hanchong and Mao, Yuchen and Hu, Wenjing and others},
  booktitle={Advances in Neural Information Processing Systems},
  volume={37},
  year={2024}
}

@article{wang2025causalcopilot,
  title={{Causal-Copilot}: An Autonomous Causal Analysis Agent},
  author={Wang, Xinyue and Zhou, Kun and Wu, Wenyi and Singh, Har Simrat and Nan, Fang and Jin, Songyao and Philip, Aryan and Patnaik, Saloni and Zhu, Hou and Singh, Shivam and others},
  journal={arXiv preprint arXiv:2504.13263},
  year={2025},
  url={https://arxiv.org/abs/2504.13263}
}

@article{xu2026agentskills,
  title={Agent Skills for Large Language Models: Architecture, Acquisition, Security, and the Path Forward},
  author={Xu, Renjun and Yan, Yang},
  journal={arXiv preprint arXiv:2602.12430},
  year={2026}
}

@article{luo2026dataagents,
  title={Data Agents: Levels, State of the Art, and Open Problems},
  author={Luo, Yuyu and Li, Guoliang and Fan, Ju and Tang, Nan},
  journal={SIGMOD 2026 Tutorial; arXiv preprint arXiv:2602.04261},
  year={2026}
}

@article{cook1977influential,
  title={Detection of Influential Observation in Linear Regression},
  author={Cook, R. Dennis},
  journal={Technometrics},
  volume={19},
  number={1},
  pages={15--18},
  year={1977},
  doi={10.1080/00401706.1977.10489493}
}

@article{huber1964robust,
  title={Robust Estimation of a Location Parameter},
  author={Huber, Peter J.},
  journal={The Annals of Mathematical Statistics},
  volume={35},
  number={1},
  pages={73--101},
  year={1964},
  doi={10.1214/aoms/1177703732}
}

@inproceedings{ng2002spectral,
  title={On Spectral Clustering: Analysis and an Algorithm},
  author={Ng, Andrew Y. and Jordan, Michael I. and Weiss, Yair},
  booktitle={Advances in Neural Information Processing Systems},
  volume={14},
  year={2001}
}

@article{rousseeuw1987silhouettes,
  title={Silhouettes: A Graphical Aid to the Interpretation and Validation of Cluster Analysis},
  author={Rousseeuw, Peter J.},
  journal={Journal of Computational and Applied Mathematics},
  volume={20},
  pages={53--65},
  year={1987},
  doi={10.1016/0377-0427(87)90125-7}
}

@misc{vedavyasvUSAHousing,
  author       = {{Vedavyasv}},
  title        = {{USA Housing}},
  howpublished = {Kaggle dataset, \url{https://www.kaggle.com/datasets/vedavyasv/usa-housing}},
  note         = {Accessed 2026-09-21}
}

\clearpage
\appendix

\section{Frozen Protocol Details}
\label{app:protocol}

Each primary condition uses seeds 0--19. The confirmation repeats the four generators at seeds 20--29 with the locked selective router. Table~\ref{tab:routing-conditions} lists the fixed preferences and expected methods.

\begin{table}[htbp]
\centering
\small
\caption{Controlled routing conditions. Each row describes one dataset per seed.}
\label{tab:routing-conditions}
\begin{tabular}{llrll}
\toprule
Condition & Generator & Rows & Preference & Expected method \\
\midrule
R-\accept{} & Linear Gaussian & 1,000 & OLS & OLS \\
R-\override{} & Paired contaminated linear & 1,000 & OLS & Huber regression \\
C-\accept{} & Gaussian blobs & 900 & $K$-means & $K$-means \\
C-\override{} & Disc and two rings & 1,000 & $K$-means & Spectral clustering \\
\bottomrule
\end{tabular}
\end{table}

For R-\accept{}, the five predictors and the response noise are independent standard normal draws. The intercept is 5 and the coefficients are $(3,-2,1.5,0,0.5)$. R-\override{} copies R-\accept{} at the same seed, replacing its first two predictor rows with $(8,-8,8,8,-8)$ and its negation and setting their response residuals to $+40$ and $-40$, respectively. The outer regression split is 80\% training and 20\% holdout, seeded by the condition seed and shared across policies. Regression diagnostics and model selection use training data.

For C-\accept{}, the three isotropic Gaussian groups have centers $(-4,-4)$, $(0,4)$, and $(4,-4)$, 300 points each, and standard deviation 0.55. C-\override{} has 250 points uniform in a disc of radius 1 and two rings of radii 3 and 5 with 250 and 500 points, respectively. Angles are uniform and ring radii have independent Gaussian noise with standard deviation 0.08. True coefficients and cluster labels are used only for post-execution scoring.

On the 80 primary instances, discordant counts were 39 in favor of the selective router and 2 in favor of no-override or constant-default, and 38 in favor of the selective router and 1 in favor of constant-complex. These are descriptive counts. The two regression conditions share the same base data within a seed.

\paragraph{Routing thresholds.}

For a training partition of $n$ rows and $d$ predictors, the default influence cutoffs are Cook's distance $4/n$, leverage $2(d+1)/n$, and absolute externally studentized residual 3. A row is high priority if Cook's distance exceeds ten times its cutoff or all three screens trigger. The relative coefficient-shift cutoff is 0.02. The near-best validation RMSE ratio for a robust candidate is 1.01.

For clustering, the local topology check constructs undirected, symmetrized nearest-neighbor connectivity graphs on standardized features for neighbor counts from 6 through 20 at these sample sizes. Candidate partitions have 2--9 connected components, each containing at least 5\% of the points. The same component count and an adjacent-partition ARI of at least 0.999999 must persist across at least $\max(3,\lceil0.30S\rceil)$ consecutive scales, where $S$ is the grid size. The first persistent range supplies the partition at its middle scale. The local graph triggers a spectral recommendation when its labels and the $K$-means labels have ARI below 0.95. The checker can also recommend spectral clustering via a separate graph-cut test when the smallest spectral component contains at least 5\% of the points and the spectral normalized cut is below 0.85 times the $K$-means cut. These values are routing heuristics.

\begin{table}[htbp]
\centering
\small
\caption{Prespecified regression-threshold sensitivity. The multiplier scales the three influence cutoffs together. The coefficient-shift threshold is fixed.}
\label{tab:thresholds}
\begin{tabular}{lrrrr}
\toprule
Multiplier & Correct & R-\accept{} & R-\override{} & Median RMSE \\
\midrule
0.50 & 20/40 & 0/20 & 20/20 & 1.013 \\
0.75 & 20/40 & 0/20 & 20/20 & 1.013 \\
1.00 & 37/40 & 18/20 & 19/20 & 1.013 \\
1.25 & 38/40 & 19/20 & 19/20 & 1.013 \\
1.50 & 38/40 & 19/20 & 19/20 & 1.013 \\
2.00 & 38/40 & 19/20 & 19/20 & 1.013 \\
\bottomrule
\end{tabular}
\end{table}

\begin{table}[htbp]
\centering
\small
\caption{Clustering agreement-threshold sensitivity. The default 0.95 was retained; no threshold was selected from these outcomes.}
\label{tab:clustering-thresholds}
\begin{tabular}{lrrr}
\toprule
Agreement cutoff & Correct & C-\accept{} & C-\override{} \\
\midrule
0.50 & 40/40 & 20/20 & 20/20 \\
0.75 & 40/40 & 20/20 & 20/20 \\
0.90 & 40/40 & 20/20 & 20/20 \\
0.95 & 40/40 & 20/20 & 20/20 \\
0.975 & 40/40 & 20/20 & 20/20 \\
0.99 & 40/40 & 20/20 & 20/20 \\
1.00 & 40/40 & 20/20 & 20/20 \\
\bottomrule
\end{tabular}
\end{table}

\section{Language-Interface Prompt Suite}
\label{app:prompts}

We used two Meta instruction models via Workers AI. Each request started from fresh state at temperature zero.
\begin{itemize}
  \item Llama 3.3 70B Instruct FP8 Fast\\
        \nolinkurl{@cf/meta/llama-3.3-70b-instruct-fp8-fast}
  \item Llama 3.1 8B Instruct FP8\\
        \nolinkurl{@cf/meta/llama-3.1-8b-instruct-fp8}
\end{itemize}

The 25-prompt language-interface suite contains the following five regression templates instantiated on R-\accept{} and R-\override{}, producing ten IDs indexed P1--P5. Every instance expected task \texttt{regression}, target \texttt{target}, features \texttt{x1}--\texttt{x5}, an OLS skill preference, and no clarification.

\begin{enumerate}[label=R\arabic*.,leftmargin=3em]
\item ``Model target from all other columns. Explain predictor associations, check assumptions, and report held-out predictive performance.''
\item ``Which variables are associated with target? Fit an appropriate regression, validate it on unseen data, and summarize uncertainty.''
\item ``Analyze target as a continuous outcome using every predictor. Diagnose the fitted model before choosing the final estimator and include visualizations.''
\item ``Build and validate a regression for target. Report predictor estimates, predictive performance, diagnostic evidence, and limitations.''
\item ``Use the available features to explain and predict target. Check regression assumptions, select a supported method, and provide a concise audit trail.''
\end{enumerate}

Each clustering template was likewise instantiated on C-\accept{} and C-\override{}, producing ten IDs. Every instance expected task \texttt{clustering}, no target, features \texttt{x1,x2}, a $K$-means skill preference, and no clarification.

\begin{enumerate}[label=C\arabic*.,leftmargin=3em]
\item ``Cluster the rows using x1 and x2, determine an appropriate structure from the data, validate stability, and visualize the result.''
\item ``Find meaningful groups from both numeric columns. Choose a suitable clustering procedure, check the result, and show the partition.''
\item ``Perform unsupervised grouping with all available features. Diagnose the data geometry, select the method, and report cluster sizes and validation evidence.''
\item ``Explore whether the observations form stable clusters in x1 and x2. Execute an appropriate algorithm and visualize the final labels.''
\item ``Use x1 and x2 for clustering. Compare the relevant structural evidence, choose a supported approach, and return an auditable summary with a plot.''
\end{enumerate}

The five ambiguity/guardrail prompts and frozen expected outcomes were:

\begin{enumerate}[label=A\arabic*.,leftmargin=3em]
\item ``Build a regression model from this table and explain the important predictors.'' Expected: regression with clarification because either \texttt{price} or \texttt{rent} could be the target.
\item ``Predict revenue from the remaining columns, treating sales as the target variable.'' Expected: regression with clarification because \texttt{revenue} and \texttt{sales} conflict as targets.
\item ``Analyze this dataset and tell me what matters.'' Expected: unknown task with clarification.
\item ``Predict price using income and neighborhood\_score, and explain both effects.'' Expected: regression targeting \texttt{price}, with clarification because \texttt{neighborhood\_score} is absent.
\item ``Use XGBoost to predict target from all five predictors and report held-out performance.'' Expected: regression targeting \texttt{target}; the unsupported model must be replanned to a valid method or receive a symbolic fallback, not clarification.
\end{enumerate}

Task accuracy therefore uses all 25 prompts. Target accuracy uses the ten executable regression instances plus A4 and A5 ($n=12$); feature exact match uses all 20 executable instances plus A4 and A5 ($n=22$); and plan validity uses the 21 records with a non-null planner status. Clarification metrics use all five A-prompts: four positives and A5 as the negative guardrail case. Each model--prompt pair was sampled once from fresh state at temperature zero.

Call-count and LLM-latency summaries include all 25 records per model, including two zero-call requests. A2 and A3 stopped before an LLM call, while A1 invoked the parser. Token and cost summaries use the 23 records per model with provider-reported usage. The archived suite also includes one semantic-structuring prompt outside the tabular-analysis scope of Table~\ref{tab:study-b}.

The 2026-09-15 Workers AI snapshot used USD per million-token input/output rates of $0.293/2.253$ for the 70B model and $0.152/0.287$ for the 8B model.

\section{Original Comparison Protocol}

The original comparison was run on 2026-06-28. Every system--task pair used the same uploaded CSV and exact one-shot prompt in three fresh conversations, with no follow-up message. DSFlow used seed 42 with semantic retrieval disabled. Manus 1.6 Lite, Gemini 3.1 Flash-Lite, and ChatGPT 5.5 Instant used their hosted web interfaces with file upload and the available analysis environment; product defaults were unchanged, while other user-visible settings were not recorded. Consequently, the comparison is tied to these dated product labels rather than their future versions.

The regression prompt was: ``I want to see what predictors affect the housing price. In addition, I want a visualization of the housing dataset for exploratory analysis.'' The target was \texttt{price}; ``affect'' was scored as predictive association rather than causation. The clustering prompt was: ``I want to do a clustering analysis of the data and visualize the clustering result. Please use all features in the columns.'' Its features were \texttt{x1,x2}. The historical regression filename contained the word \texttt{contaminated}, which may cue a hosted assistant; a neutral-filename replication was not performed and remains a limitation.

The adjudicated regression/clustering run scores were DSFlow $7,7,7/7,7,7$ and Manus $3,2,3/1,5,5$. Gemini scored $2,2,2/1,1,2$, and ChatGPT scored $1,1,1/2,1,2$. DSFlow's identical values are fixed-seed deterministic recomputations, not independent stochastic replicates.

\section{Component Counterfactuals}

\begin{table}[htbp]
\centering
\small
\caption{Component configurations evaluated on the two original cases.}
\label{tab:components}
\begin{tabular}{p{0.27\textwidth}p{0.30\textwidth}p{0.33\textwidth}}
\toprule
Configuration & Regression & Clustering \\
\midrule
Full deterministic core & Robust; RMSE 100,131 & Spectral; sizes 250/250/500 \\
Oracle plan & Same as full core & Same as full core \\
Constant complex & Same as full core & Same as full core \\
Override disabled & OLS; RMSE 127,139 & $K$-means; six sectors \\
Execution guardrail off & Same as full core & Same as full core \\
LLM only & Failed before execution & $K$-means; six groups \\
\bottomrule
\end{tabular}
\end{table}

\section{Inter-Rater Agreement}

One author rater and one independent computer-science graduate student independently scored the 24 original comparison runs. Across 168 binary judgments they agreed on 153 (91.1\%, Cohen's $\kappa=0.82$) before adjudication. Blinding removed explicit product identifiers but not all stylistic cues. The 15 disagreements were then jointly adjudicated (13 resolved to the author rating and 2 to the independent rating); Table~\ref{tab:original-comparison} uses these adjudicated scores. Table~\ref{tab:irr} reports agreement by criterion. The lowest-agreement criterion was generic influence detection ($\kappa=0.08$), reflecting disagreement over whether routine outlier removal constitutes an influence diagnostic.

\begin{table}[htbp]
\centering
\small
\begin{tabular}{lcc}
\toprule
Rubric criterion & Agreement & Cohen's $\kappa$ \\
\midrule
Regression: detects influence & $50.0\%$ & $0.08$ \\
Regression: identifies injected rows & $83.3\%$ & $0.57$ \\
Regression: handles contamination & $83.3\%$ & $0.64$ \\
Regression: predictive validation & $100\%$ & $1.00$ \\
Regression: quantifies associations & $100\%$ & $1.00$ \\
Regression: interprets limits & $100\%$ & $1.00$ \\
Regression: visualization & $100\%$ & N/A \\
Clustering: selects three groups & $100\%$ & $1.00$ \\
Clustering: recovers geometry & $100\%$ & $1.00$ \\
Clustering: non-convex method & $100\%$ & $1.00$ \\
Clustering: reports sizes & $100\%$ & $1.00$ \\
Clustering: validates structure & $75.0\%$ & $0.50$ \\
Clustering: visualization & $83.3\%$ & $0.64$ \\
Clustering: avoids causal claims & $100\%$ & N/A \\
\midrule
Overall ($168$ judgments) & $91.1\%$ & $0.82$ \\
\bottomrule
\end{tabular}
\caption{Inter-rater agreement before adjudication. Each criterion covers 12 runs and the overall row covers 168 binary judgments. Cohen's kappa is undefined when both raters assign the same single category to every run.}
\label{tab:irr}
\end{table}

\section{Evidence Lineage}

The experiment archive contains the final selective-routing results (77/80), the additional-seed confirmation (39/40), the 52 archived language-interface records (50 reported here), the component configurations, both threshold sweeps, and the DS-Agent preflight. The DS-Agent custom-task interface required supervised Kaggle-style train/test files, a submission metric, and starter training code; it exposed no task-equivalent unlabeled clustering path, so no score was assigned. All runs within each reported scope contribute to the stated denominators.

\clearpage
\section{Original-Case Visual Evidence}

The following archived figures document representative outputs from the original seed-42 regression and clustering cases. They provide interface and artifact evidence rather than additional independent runs.

\begin{appendixfigure}
\includegraphics[width=0.96\textwidth]{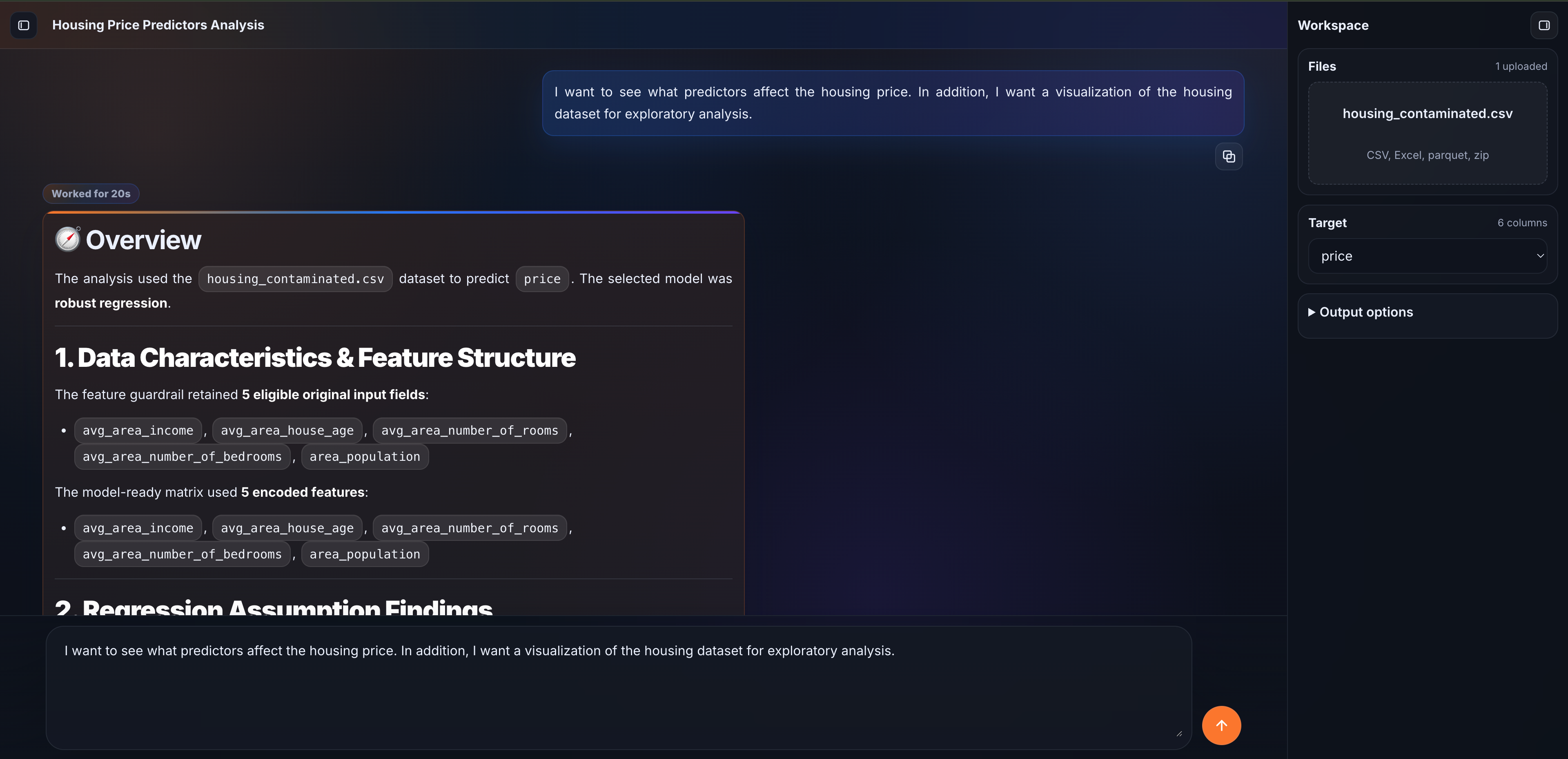}
\caption{Archived contaminated-regression interface overview showing the submitted request, uploaded dataset, target, retained feature set, and selected robust-regression method.}
\label{fig:appendix-regression-overview}
\end{appendixfigure}

\medskip
\begin{appendixfigure}
\includegraphics[width=0.82\textwidth]{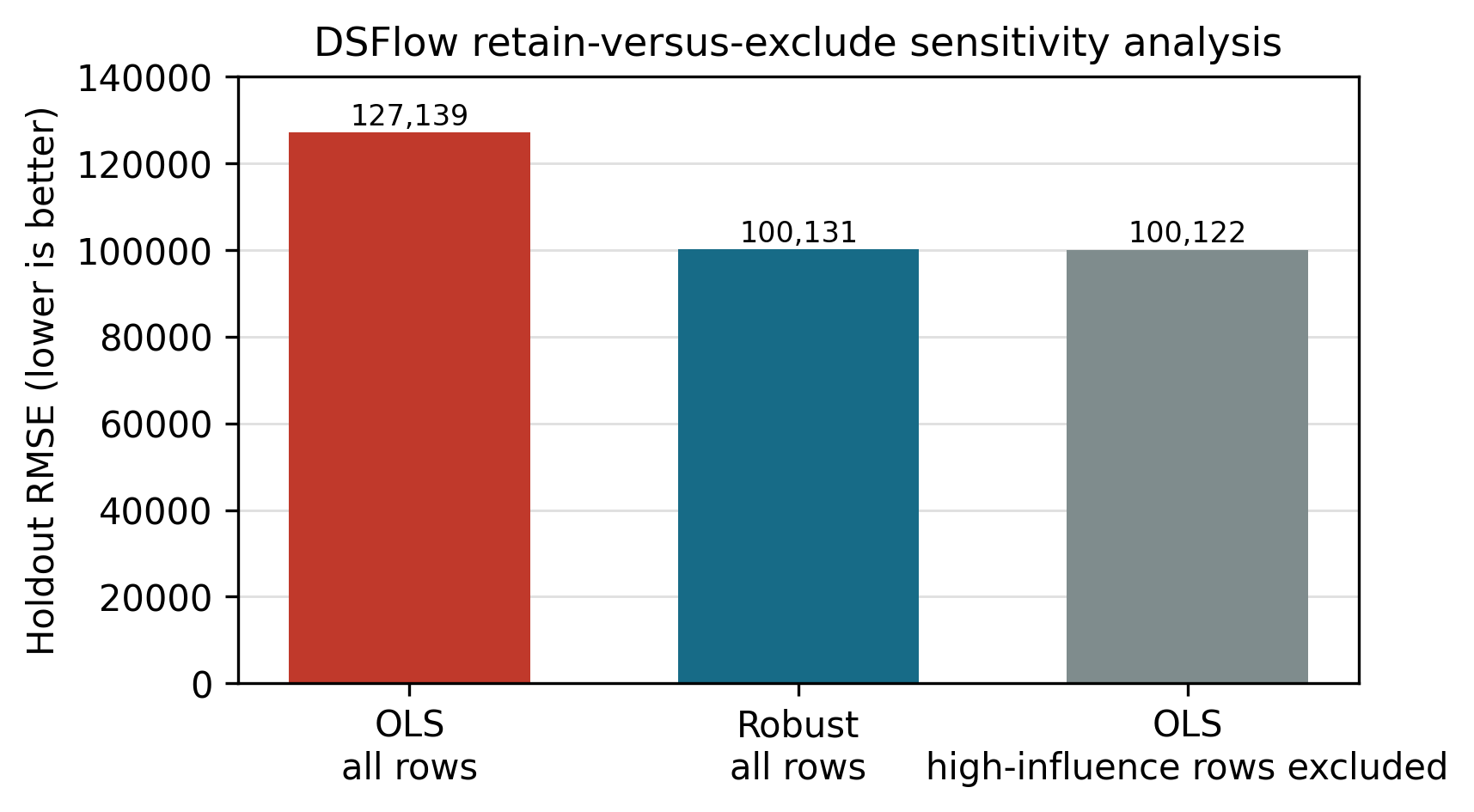}
\caption{Original contaminated-regression sensitivity analysis. Holdout RMSE was 127,139 for OLS on all training rows, 100,131 for robust regression retaining all rows, and 100,122 for OLS after excluding the three highest-influence training rows.}
\label{fig:appendix-regression-sensitivity}
\end{appendixfigure}

\clearpage
\begin{appendixfigure}
\begin{minipage}[t]{0.485\textwidth}
\centering
\includegraphics[width=\linewidth]{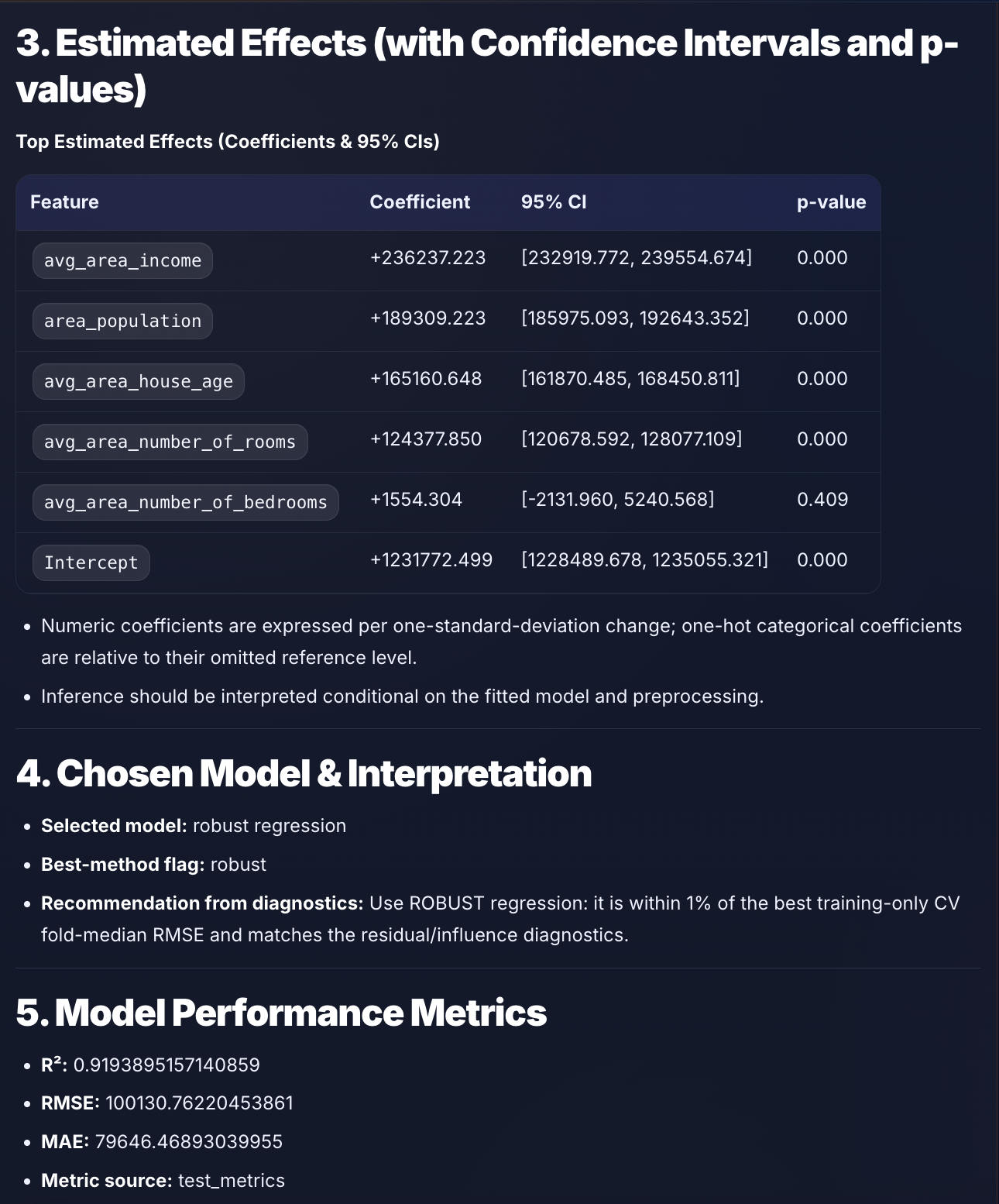}
\small (a) Coefficient intervals, selected method, and held-out metrics.
\end{minipage}\hfill
\begin{minipage}[t]{0.485\textwidth}
\centering
\includegraphics[width=\linewidth]{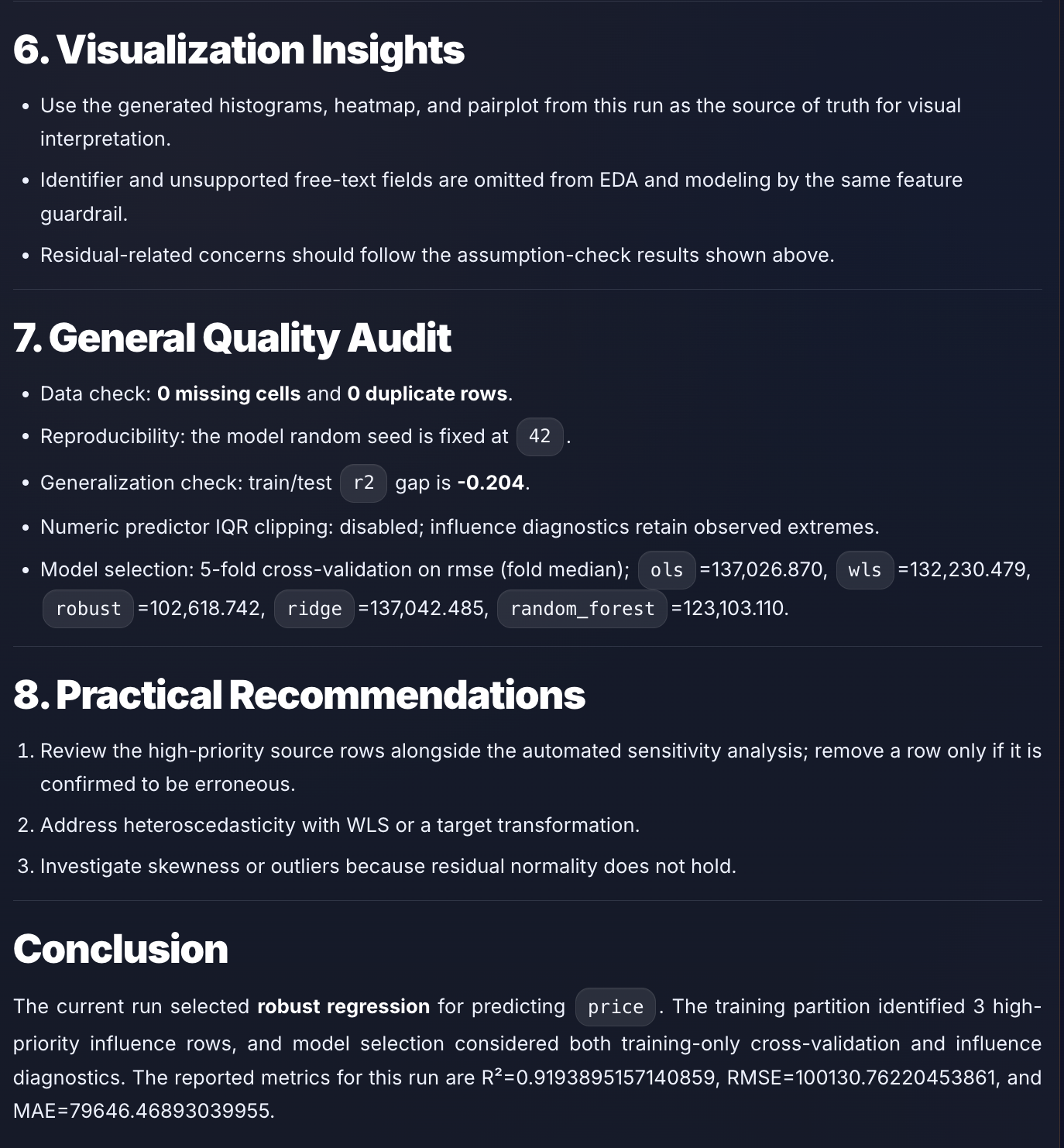}
\small (b) Quality audit, recommendations, and run conclusion.
\end{minipage}
\caption{Archived rendered sections of the original contaminated-regression report. Coefficients are conditional model estimates rather than causal effects.}
\label{fig:appendix-regression-report}
\end{appendixfigure}

\medskip
\begin{appendixfigure}
\begin{minipage}[t]{0.485\textwidth}
\centering
\includegraphics[width=\linewidth]{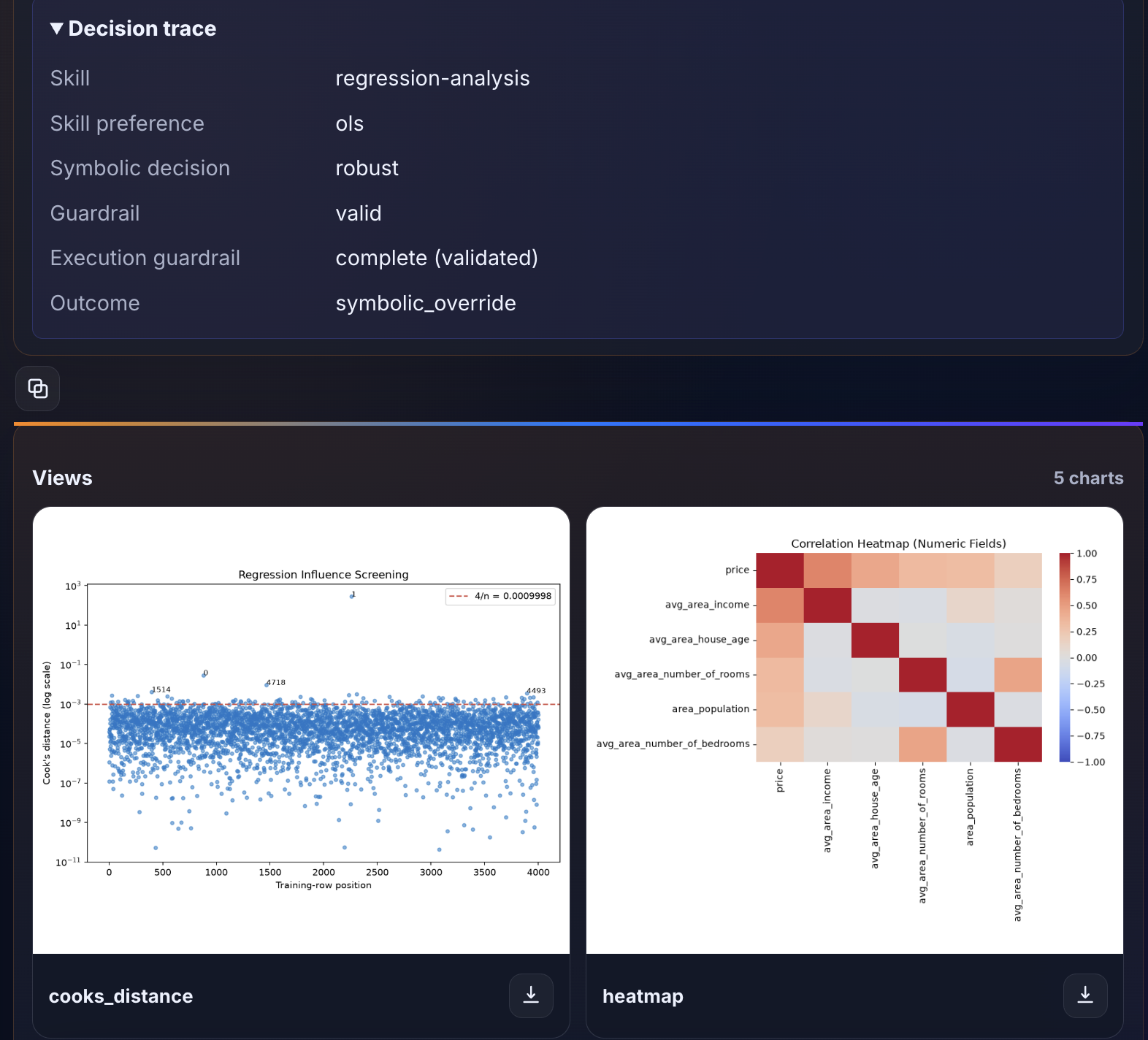}
\small (a) Preference-to-decision trace and diagnostic views.
\end{minipage}\hfill
\begin{minipage}[t]{0.485\textwidth}
\centering
\includegraphics[width=\linewidth]{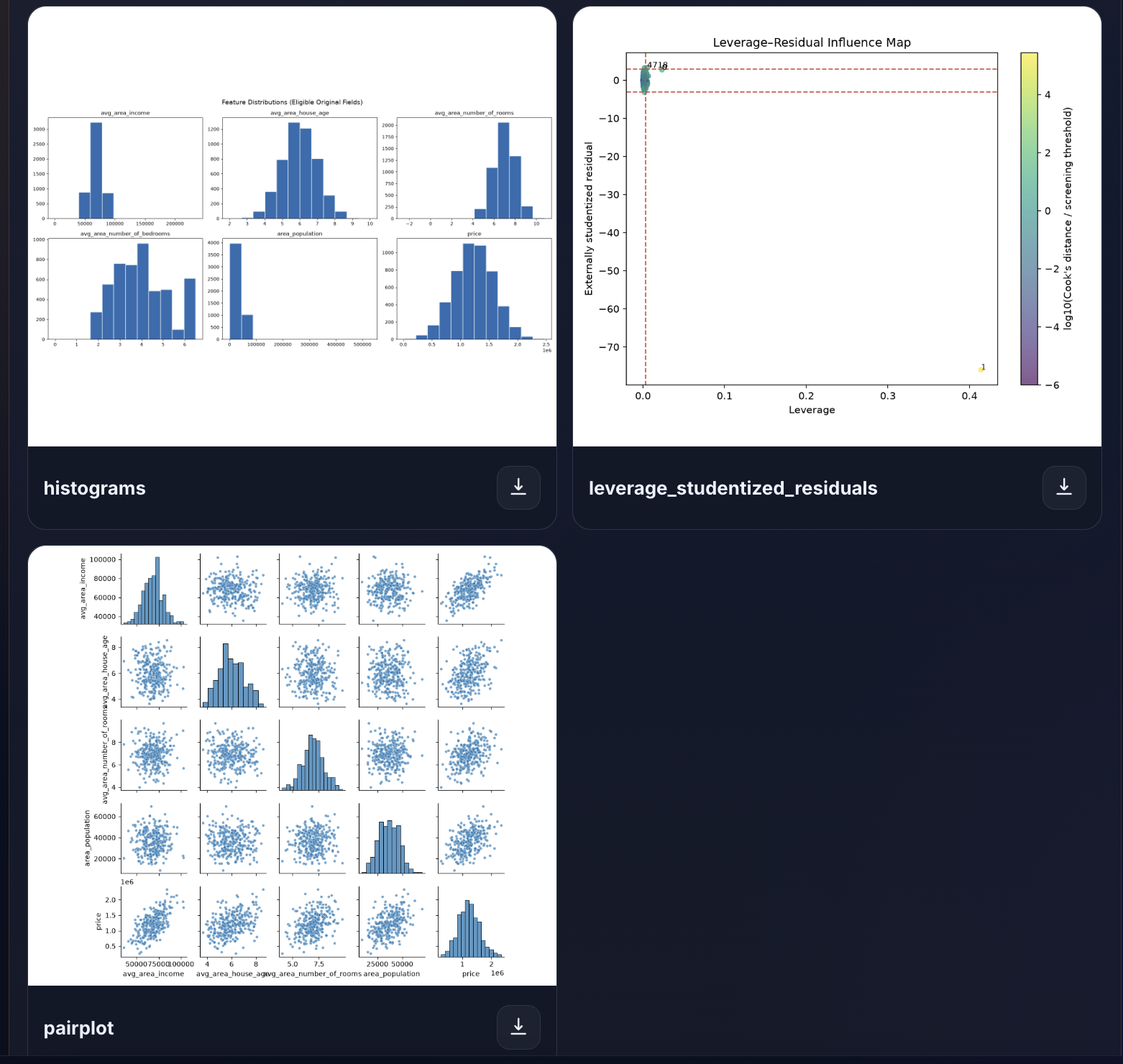}
\small (b) Distribution, leverage-residual, and pairwise views.
\end{minipage}
\caption{Archived regression decision trace and generated diagnostics. The workflow used Cook's distance, leverage, and studentized residuals to prioritize influential training rows before selecting robust regression.}
\label{fig:appendix-regression-diagnostics}
\end{appendixfigure}

\clearpage
\begin{appendixfigure}
\includegraphics[width=0.92\textwidth]{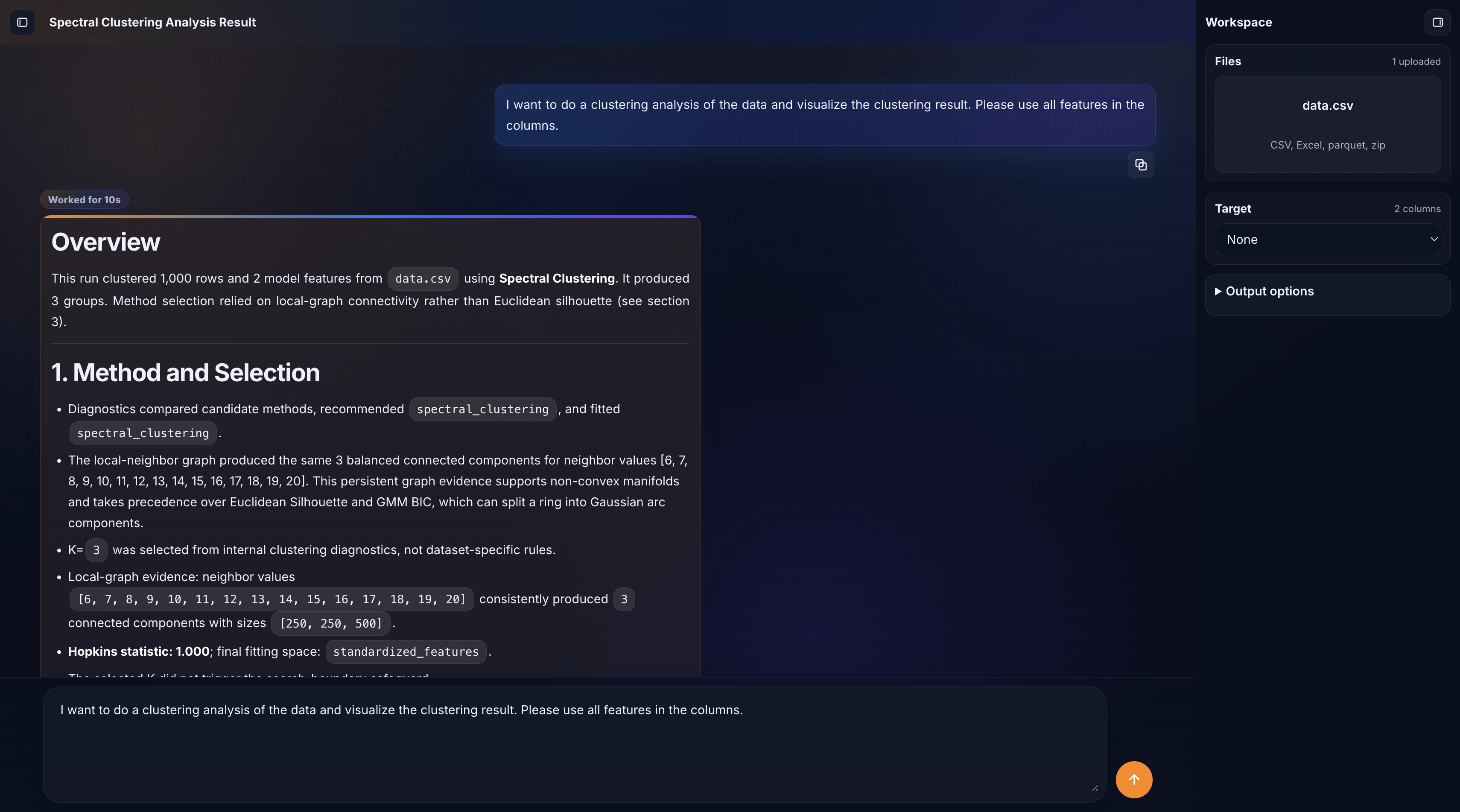}
\caption{Archived non-convex-clustering output. The local graph supported three connected components across neighbor counts 6--20, and the workflow selected spectral clustering.}
\label{fig:appendix-clustering-overview}
\end{appendixfigure}

\medskip
\begin{appendixfigure}
\begin{minipage}[t]{0.495\textwidth}
\centering
\includegraphics[width=\linewidth]{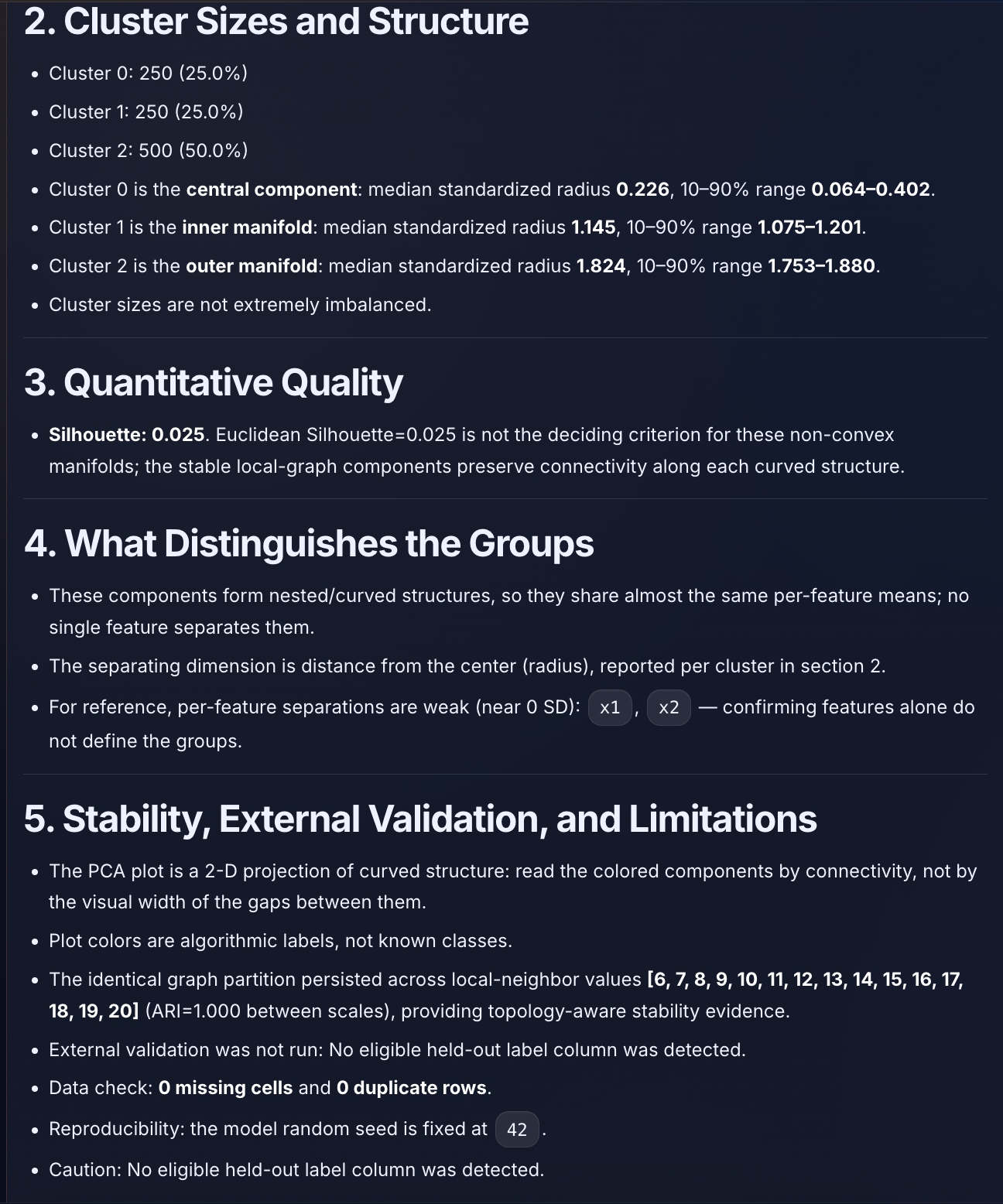}
\small (a) Component sizes, radial structure, and stability limits.
\end{minipage}\hfill
\begin{minipage}[t]{0.495\textwidth}
\centering
\includegraphics[width=\linewidth]{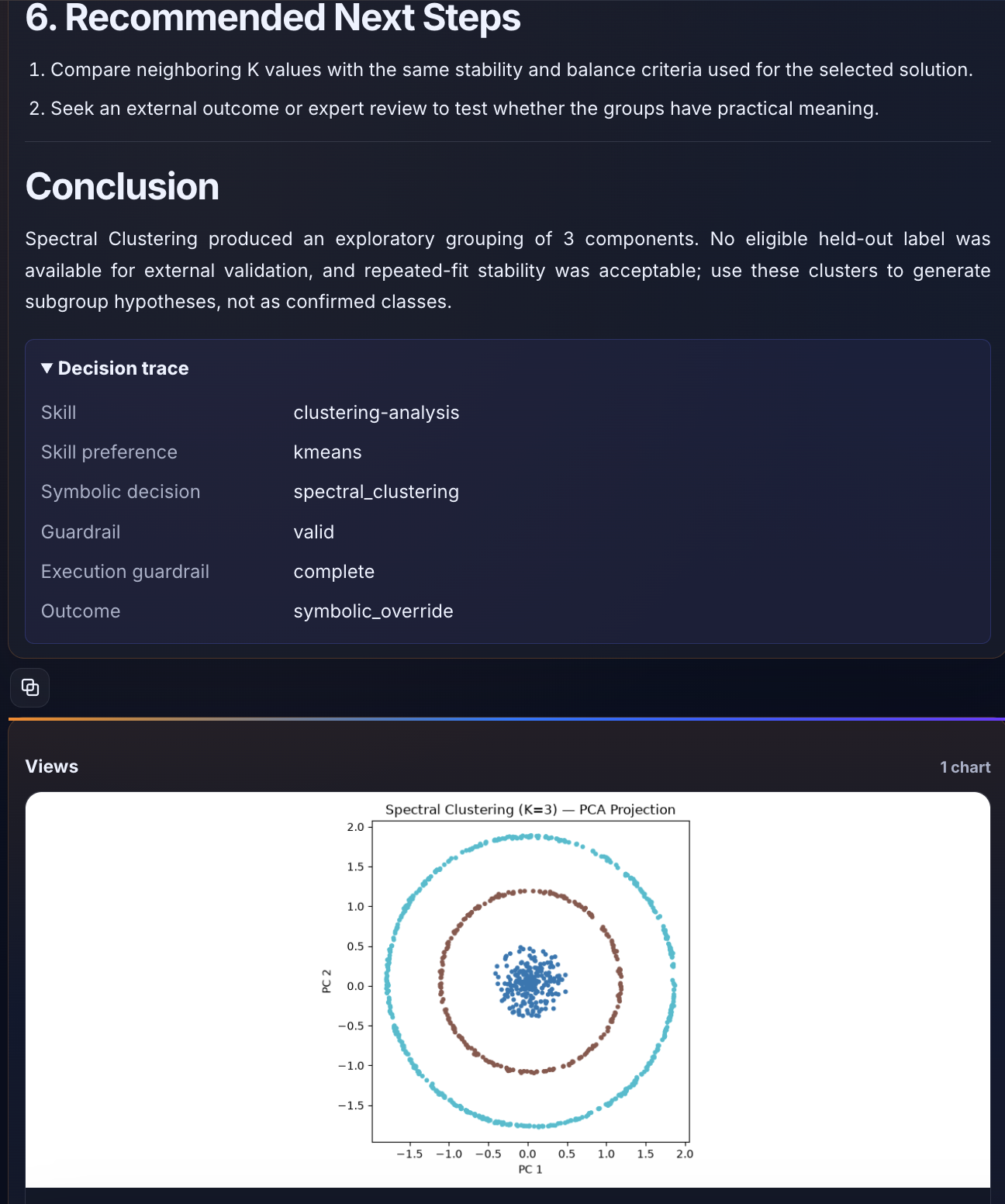}
\small (b) Recommendation, decision trace, and spectral result.
\end{minipage}
\caption{Archived clustering report, symbolic-override trace, and final visualization. The spectral partition recovered the central disc and the inner and outer rings, with 250, 250, and 500 points, respectively.}
\label{fig:appendix-clustering-report}
\end{appendixfigure}

\end{document}